# ReDWINE: A Clinical Datamart with Text Analytical Capabilities to Facilitate Rehabilitation Research

David Oniani[1], Bambang Parmanto[1], Andi Saptono[1], Allyn Bove[3], Janet Freburger[3], Shyam Visweswaran[4,5,6], Nickie Cappella[4,6], Brian McLay[4,6], Jonathan C. Silverstein[4,6], Michael J. Becich[4,6], Anthony Delitto[3], Elizabeth Skidmore[2], Yanshan Wang, PhD[1,4,5,6]

[1]Department of Health Information Management, University of Pittsburgh, Pittsburgh, PA, USA; [2]Department of Occupational Therapy, University of Pittsburgh, Pittsburgh, PA, USA; [3]Department of Physical Therapy, University of Pittsburgh, Pittsburgh, PA, USA; [4]Department of Biomedical Informatics, University of Pittsburgh School of Medicine, Pittsburgh, PA; [5]Intelligent Systems Program, University of Pittsburgh, Pittsburgh, PA; [6]Clinical and Translational Science Institute, University of Pittsburgh, Pittsburgh, PA

**Abstract**

*Rehabilitation research focuses on determining the components of a treatment intervention, the mechanism of how these components lead to recovery and rehabilitation, and ultimately the optimal intervention strategies to maximize patients' physical, psychologic, and social functioning. Traditional randomized clinical trials that study and establish new interventions face several challenges, such as high cost and time commitment. Observational studies that use existing clinical data to observe the effect of an intervention have shown several advantages over RCTs. Electronic Health Records (EHRs) have become an increasingly important resource for conducting observational studies. To support these studies, we developed a clinical research datamart, called ReDWINE (Rehabilitation Datamart With Informatics iNfrastructure for rEsearch), that transforms the rehabilitation-related EHR data collected from the UPMC health care system to the Observational Health Data Sciences and Informatics (OHDSI) Observational Medical Outcomes Partnership (OMOP) Common Data Model (CDM) to facilitate rehabilitation research. The standardized EHR data stored in ReDWINE will further reduce the time and effort required by investigators to pool, harmonize, clean, and analyze data from multiple sources, leading to more robust and comprehensive research findings. ReDWINE also includes deployment of data visualization and data analytics tools to facilitate cohort definition and clinical data analysis. These include among others the Open Health Natural Language Processing (OHNLP) toolkit, a high-throughput NLP pipeline, to provide text analytical capabilities at scale in ReDWINE. Using this*

*comprehensive representation of patient data in ReDWINE for rehabilitation research will facilitate real-world evidence for health interventions and outcomes.*

**Introduction**

According to the World Health Organization (WHO), rehabilitation is defined as a set of interventions designed to optimize functioning and reduce disability in individuals with acute and chronic diseases and conditions (e.g., stroke, chronic pain, surgery, cancer) as they interact with their environment" [1]. Researchers in rehabilitation have used randomized clinical trials (RCTs), the gold standard for evidence-based medicine, to study mechanisms of change and to provide evidence for the efficacy of rehabilitation interventions. However, the use of randomized controlled trials (RCTs) in rehabilitation research presents several challenges, including lack of generalizability of the findings, high cost, and high time commitment [2].

Observational studies have been used to observe the effects of interventions in real world settings without experimental assignment. Observational studies have several advantages over RCTs, including greater external validity and being faster and less expensive to conduct. With the emergence of electronic health records (EHRs) in major health care systems, EHRs have become an increasingly important resource for conducting observational studies [3]. Having access to EHR data from large health care systems is particularly useful for rehabilitation research since many patients who are candidates for rehabilitation receive it across multiple settings. For example, an individual admitted to the hospital for a stroke may start rehabilitation in the acute care hospital but then may be transferred to an inpatient rehabilitation facility or skilled nursing facility to continue rehabilitation, followed by a transfer to the community where s/he may receive rehabilitation in the home, the outpatient setting, or both. Accessing EHR data from these various settings would provide a complete picture of the patient's rehabilitation journey and provide a more accurate picture of the value and effectiveness of rehabilitation.

To facilitate EHR-based observational studies, the Observational Health Data Sciences and Informatics (OHDSI) community has developed open-source standardization vocabulary and software tools, such as the Observational Medical Outcomes Partnership Common Data Model (OMOP CDM). The OMOP CDM has established a standardized method for mapping and integrating EHR data, as well as many other types of data (e.g., claims data, transactional data), into a universal format of data linking patients, measurements, observations, visits, procedures, and so on. In addition, the OHDSI community also offers open-source data analytics tools to facilitate observational research. ATLAS [4] is a web-based tool that provides a user interface to visualize data and standardized vocabulary and to define cohorts on the observational data in the OMOP CDM format. HADES [4] is a data analytics tool that provides a set of R packages for large scale analytics, including population characterization, population-level causal effect estimation, and patient-level prediction.

Unstructured narratives in the EHR, such as clinical notes and procedure notes, are critical in rehabilitation research. Notes provide a wealth of information about patients' medical histories, symptoms, treatments, and lifestyle factors that structured EHRs alone cannot capture [5]. Some estimates on how much research valued data are contained in clinical notes range from 70 to 80 percent or higher [6]. For example, in a physical therapy session, exercise-specific information is only available in the clinical note. Such information is valuable for clinical researchers who are interested in studying the association of different exercises and therapy outcomes. By automating the process of extracting information from unstructured EHRs, clinical and translational researchers can save time and increase their efficiency, while also gaining a more comprehensive view of patient data. The OMOP CDM has incorporated the representations associated with unstructured EHRs by introducing NOTE and NOTE_NLP tables, which store clinical textual data and the output from clinical NLP tools, respectively. However, the OHDSI community currently does not provide clinical NLP tools to extract medical concepts from unstructured EHRs.

In this article, we describe the development of a clinical research datamart prototype, called ReDWINE (Rehabilitation Datamart With Informatics iNfrastructure for rEsearch), that transforms rehabilitation-related EHR data collected in a large health care system to the OMOP CDM to facilitate rehabilitation research. The standardized EHR data stored in RedWINE could reduce the time and effort required to harmonize and map data and enable researchers to pool and analyze data from multiple sources, leading to more robust and comprehensive research findings. We also deployed the ATLAS and HADES tools in ReDWINE to facilitate cohort definition and clinical data analysis. In addition, we adopted and deployed the Open Health Natural Language Processing (OHNLP) toolkit [7], a high-throughput NLP pipeline, to provide text analytical capabilities at scale. Using this comprehensive representation of patient data in ReDWINE for rehabilitation research could facilitate real-world evidence for health interventions and outcomes.

**Methods**

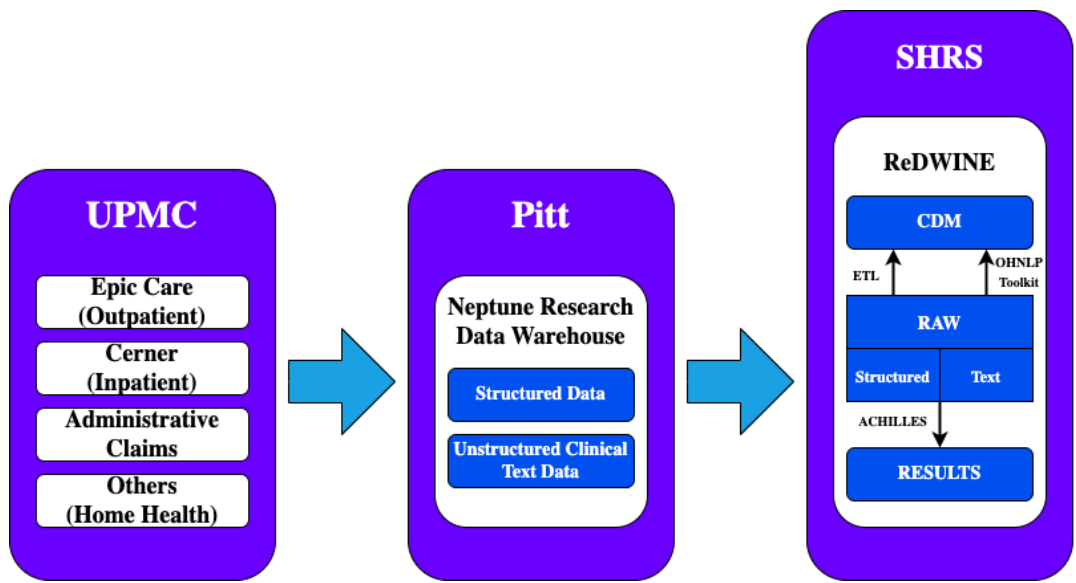

*Figure 1. ReDWINE Architecture and Data Pipeline.*

In this section, we describe the implementation details of ReDWINE. ReDWINE is designed as a self-service informatics and data analytics platform that leverages EHR data from the University of Pittsburgh Medical Center (UPMC) and supports observational rehabilitation research at the School of Health and Rehabilitation Sciences (SHRS) at the University of Pittsburgh.

*Data Collection*

The University of Pittsburgh School of Medicine's Department of Biomedical Informatics designed and implemented a research data warehouse called Neptune [8] that extracts, transforms and stores EHR data from UPMC as part of their Business Associates Agreement (BAA) with the UPMC. Neptune [8] has a data layer that contains structured data de-identified to HIPAA Limited Data with dates and unstructured EHR clinical text data for research purposes. Due to the complex nature of multiple UPMC EHR systems, EHR data are stored in Neptune with minimal filtering and processing. The ReDWINE team worked with the Neptune team and retrieved rehabilitation-related EHR data from the data layer and transformed them into the OMOP CDM for rehabilitation researchers.

As the first step in developing the ReDWINE prototype, we used EHR data from a cohort of patients diagnosed with stroke. The reason we selected stroke is that it is a prevalent disease and many rehabilitation researchers at SHRS focus on rehabilitation therapies post-stroke. We defined a cohort of patients diagnosed with stroke and admitted to a UPMC acute care hospital (see the supplemental file for ICD-10 codes) between January 1, 2016 and December 31, 2016 at UPMC. We collected any available EHR data on these patients for the 12 months following the stroke across the outpatient settings, including demographics, diagnosis billing codes, procedures codes, encounter notes, procedure notes, which were created between January 1, 2016 and December 31, 2018. The University of Pittsburgh's Institutional Review Board (IRB) reviewed and approved development of ReDWINE  (IRB #21040204).

*Database*

We used the PostgreSQL[1] database in our implementation for RedWINE since it is well supported by the OHDSI community. PostgreSQL is a free and open-source object-relational database management system (ORDBMS) focused on extensibility and SQL compliance. In the database, we defined three schemas, namely RAW, CDM, and RESULTS. Every schema has its own purpose and contains corresponding tables:

- RAW schema: As the name suggests, the RAW schema holds the raw data obtained from Neptune that is subsequently transformed, standardized, and mapped via the Extract, Transform, Load (ETL) process[2]. This schema is based on the EHR data received from Neptune and thus can be flexible in its design. The tables in this schema depend on the nature and modality of the data.
- CDM schema: The CDM schema contains standardized EHR data in OMOP format following the standardized vocabulary and tables defined in the OMOP CDM.
- RESULTS schema: The RESULTS schema contains information and statistics generated by the Automated Characterization of Health Information at Large-scale Longitudinal Evidence Systems (ACHILLES) tool[3], which validates data quality and determines whether data meet a given requirement.

Implementation details about each schema in ReDWINE, the ETL process, the text analytical tool can be found in the supplemental file due to the word limit of the manuscript.

*Data Quality Assurance*

Because the data transformed between different standards and codes, we conducted data quality assurance (QA) analysis to assess the quality of the ETL scripts and to make sure that there is no significant data loss in the pipeline. This is done by computing record counts of the original data in the RAW schema and in the CDM schema followed by generating statistics to assess data loss. The step ensures that the pipeline is robust and can handle future streams of data well.

---

[1] https://www.postgresql.org/
[2] https://ohdsi.github.io/TheBookOfOhdsi/ExtractTransformLoad.html
[3] https://github.com/OHDSI/Achilles

**Results**

*Statistics of ReDWINE Prototype*

Table 1 lists the statistics of the ReDWINE prototype using the stroke cohort. Currently there are a total of 13,604 patients in the database with 6,673 males and 6,931 females. There are ~2 million diagnoses entries for the cohort with an average of 140 diagnoses per person. ReDWINE contains ~2 million clinical notes for the cohort and a total of ~304 million medical concepts after the NLP processing, with an average of 22,000 medical concepts per patient.

**Table 1. Cohort statistics of the ReDWINE prototype.**

|  | Number |
|---|---|
| **Patients** | 13,604 |
| Male | 6,673 |
| Female | 6,931 |
| **Diagnoses** | 1,909,175 |
| Diagnoses per person (average) | 140 |
| **Visits** | 1,987,791 |
| Visits per person (average) | 146 |
| **Procedures** | 1,214,388 |
| Procedures per person (average) | 89 |
| **Notes** | 2,096,415 |
| Notes per person (average) | 154 |
| **Clinical NLP Concepts (Note_NLP table)** | 304,228,932 (60,627 unique) |
| Concepts per person (average) | 5,018 |

*Data QA Results*

Table 2 shows the results of data QA analysis on records counts for the RAW and CDM schemas. We found that the data transformation ETL scripts delivered good performance for demographics, diagnosis, encounter visits, and procedures with minimal data loss (ranges from 0% to 9.627%). For the RAW.demographics vs CDM.person comparison, there was no data loss and every single person was mapped from the table in the RAW schema to the corresponding table in the CDM schema. As for RAW.diagnoses vs CDM.condition_occurrence, there were 2,475 records left unmapped (0.129%). For RAW.encounters vs CDM.visit_occurence, all of the records were successfully mapped to the CDM table without data loss. Finally, for RAW.procedures vs CDM.procecure_occurrence, we were not able to map 129,371 records since a number of CPT-4 codes that were not present in the OMOP CDM standardized vocabulary had to be excluded. This has resulted in the exclusion of 9.627% of the original 1,343,759 records. Overall, the record counts are consistent between the RAW and CDM schemas, which indicates the data were successfully transferred to the OMOP CDM.

**Table 2. Results of data quality assurance analysis for ReDWINE prototype.**

| Comparison | RAW | CDM | Difference | RAW Lost (%) |
|---|---|---|---|---|
| RAW.demographics vs CDM.person | 13,604 | 13,604 | 0 | 0 |
| RAW.diagnoses vs CDM.condition_occurrence | 1,911,650 | 1,909,175 | 2,475 | 0.129 |
| RAW.encounters vs CDM.visit_occurrence | 1,987,791 | 1,987,791 | 0 | 0 |
| RAW.procedures vs CDM.procedure_occurrence | 1,343,759 | 1,214,388 | 129,371 | 9.627 |

*Case Study 1: Use ATLAS to define cohorts and visualize results*

Suppose a rehabilitation researcher is interested in identifying a cohort of patients who had physical therapeis and were diagnosed with type 2 diabetes followed by stroke, he/she could define these criteria using the cohort definition tool in ATLAS. The researcher could use the OHDSI Athena tool to identify the standardized OMOP CDM concept IDs to define three concept sets for the inclusion criteria. Using the defined concept sets, the researcher could define the cohort in ATLAS as shown in Figure 2. The cohort entry event is defined as the cohort index time when subjects in the database are entered into the cohort. In our case, we used the stroke diagnosis (concept ID: 35207821) as the entry event. Then the researcher could define inclusion criteria to further restrict the cohort. In this case study, we used physical therapy (concept ID: 2314284) and type 2 diabetes (concept ID: 35206882). One could add more related concept IDs in each concept set, though we only used one concept ID for demonstration purposes. Then the researcher could generate the cohort in the Generation function. In this case study, we could identify five patients who met the cohort criteria.

**Figure 2.** Cohort definition inside Atlas.

*Case Study 2: Use NLP-extracted concepts to identify patients*

Clinical information within unstructured EHRs can be very important for clinical and translational researchers. The NLP-extracted concepts in the ReDWINE could promote the use of such information for observational studies. In a case study, we assume that a researcher wanted to identify patients who experienced one or more falls. We used the Note_NLP table and the note_nlp_concept_id "436583" to identify the patients who had fall events mentioned in the clinical notes. We identified 10,322 patients using the HADES tool. We loaded these patients' data into an R dataframe for further analysis. We also found that in the Note_NLP table there were 358 unique lexical variants of the fall event as documented in clinical notes. Examples of different variants include "accident &fell", "Accident a fall", "accident and fall", "accident and falling", "accident and fell", "accident from falling", "accident/ fall", "accidents/falls", "down Falls", "down Fall", "down and falling", "down and falls", "fall", and "fall and injured". The R script to identify these patients can be found in the supplemental file. In other study [9], we leveraged the OHNLP toolkit in ReDWINE and developed a customized rule-based NLP algorithm to extract physical rehabilitation exercise information from clinical notes. The physical rehabilitation exercise information included 101 medical concepts across nine categories, including type of motion, side of body, location on body, plane of motion, duration, set and rep information, exercise purpose, exercise type, and body position. The text analytical capabilities on EHRs offered by ReDWINE represent a significant advancement towards realizing the commitment to precision rehabilitation [10].

**Discussion**

Given the availability of increasing volumes of EHR data and improving computing power, the ability to access and utilize near real-time data for both research and clinical purposes has become a reality. We present an approach that can be used to harmonize and standardize EHR data from multiple settings and that uses different EHR software (e.g., Epic, Cerner) using rehabilitation care for stroke as a use case example.  precision rehabilitation has recently gained interest in both research and practice that seeks to deliver the precise and individualized intervention at the right time [11]. The information technology infrastructure, standardized data elements, common data model, and data analytical platforms developed in ReDWINE and observational studies enabled by ReDWINE could be leveraged to advance precision rehabilitation research and ultimately to maximize function and minimize disability in patients with acute illness or injury and chronic conditions. Furthermore, ReDWINE offers a platform for learning health systems research that seeks to link health system data and experiences with external evidence to improve practice. As a result, ReDWINE supports a system where patients get higher quality, safer, and more efficient care.

The ability to follow patients over time and across settings is particularly relevant for rehabilitation research since rehabilitation care often occurs across settings and over several months. This feature, however, has larger clinical and health system implications when it is useful to understand patient outcomes and health care utilization over time. For example, with payment models such as bundled payment and accountable care organizations, health systems receive a lump sum payment to manage patients over a specific period of time. Understanding the patient's healthcare utilization trajectory over that time period can provide useful information to improve the quality and value of care.

The inclusion of NLP tools in ReDWINE enables rehabilitation researchers to study the effects of specific interventions to a much greater extent than was possible without NLP tools. Specific exercises and other interventions provided by rehabilitation providers are often only found within a free-text clinical note within a patient's record. Therefore, simply examining CPT codes billed during a visit typically cannot provide the level of detail needed for a researcher to ascertain the effects of a specific intervention. For example, a physical therapist may bill for three 15-minute units of "therapeutic exercise" during a session. Using only structured data fields, it is typically not possible to know what exercises were performed. However, with NLP tools, researchers can extract data regarding the specific exercises and dosage (sets, repetitions, frequency, duration) provided, as shown in [9]. This will enable researchers to be much more precise in determining optimal treatment strategies for individuals seeking rehabilitation care.

The feature that ReDWINE leverages the OMOP CDM and the standardized vocabulary is critical for the rehabilitation EHR data because UPMC uses multiple EHR systems from different vendors across its various hospitals [8]. For example, different UPMC Home Health services use different

EHR systems. Although UPMC is moving to an integrated Epic EHR system, integrating historical EHR data from multiple systems is still critical for rehabilitation research. ReDWINE is able to integrate and harmonize EHR data from multiple sources that follow different data formats and coding systems, and resolve any discrepancies between sources to ensure consistency and accuracy. It provides a flexible and scalable framework, enabling rehabilitation researchers to conduct observational studies, health services research, and quality improvement, as well as facilitating multi-site studies.

Although ReDWINE is primarily developed for observational studies and health services research, its potential of storing comprehensive EHR data could also be used to assess RCT feasibility [12] as well as to enhance recruitment of underrepresented populations in the local health system. As is well known, the low recruitment rate of underrepresented populations of racial and ethnic minorities in clinical trials remains a problem and has led to many health disparities. Using rich demographic information in ReDWINE from multiple sources has the potential to help identify underrepresented populations for RCTs.

*Limitations*

There are several limitations to ReDWINE. First, ReDWINE is still a prototype implemented using a small size of EHR dataset from a single data source. Hence, the system was not analyzed in the production context, with many users and a continuous stream of data. That being said, we have implemented all the necessary components and functions in ReDWINE to be able to store more data. Second, because we wanted to make the study accessible and reproducible, for all major data analysis and exploration tools, we relied on OHDSI framework and tools (e.g., ATLAS, HADES). Thus, the development and usage of custom, non-OHDSI tools was not thoroughly explored. Finally, while OMOP CDM has several advantages for observational studies, it is not the only standardized CDM. Thus, ReDWINE may also have limitations that existed in the OMOP CDM.

*Future Work*

*We have* built a ReDWINE prototype using EHR data from a cohort of patients with stroke. The prototype has necessary functions to facilitate rehabilitation research. It is in its early stage to serve multiple users for research at the University of Pittsburgh SHRS. An immediate action item is to have several test users utilize ReDWINE for their observational studies and test the cohort definition and data analytics functions and seek their feedback to improve the prototype. To make ReDWINE a self-service tool for researchers, there are several items to be considered. First, we will work with the institution IRB to ensure that the investigator has proper IRB approvals prior to using ReDWINE. We will implement auditing and privacy settings to ensure the system is being used to its full potential by investigators. Second, we will develop a service and maintenance business model to ensure that the support team of ReDWINE is sustainable. Third, we will collaborate with the Neptune team to ensure the EHR data can be updated regularly. Fourth, we will develop several R wrappers for frequently used data retrieval functions for the HADES tool. This will make the data analytics tool more convenient for researchers without SQL experience. Finally, with the rapid emergence of Artificial Intelligence (AI) models and techniques, it can be interesting to incorporate such capabilities into the data analytics tool in ReDWINE. While it is currently possible to automatically perform statistical analyses and training AI models (e.g., AutoML [13]), such functions have not been integrated into ReDWINE. Thus, implementing a pipeline for AI model training could be another avenue of exploration in the future work.

**Authors' contributions**

DO: designed; wrote the manuscript, BP: conceptualized the study; edited the manuscript, AS: edited the manuscript, AB: edited the manuscript, JF: edited the manuscript, SV: conceptualized the study; edited the manuscript, NC: retrieved the data; edited the manuscript, BM: retrieved the data, JCS: conceptualized the study; edited the manuscript, MJB: conceptualized the study; edited the manuscript, AD: conceptualized the study; edited the manuscript, ES: conceptualized the study; edited the manuscript, YW: conceptualized the study; analyzed the data; wrote the manuscript. All authors read and approved the final manuscript.

**Acknowledgements**

The authors would like to acknowledge the OHDSI community for providing extensive documentation of the CDM and releasing the OMOP data analytics tools. This project was supported by the University of Pittsburgh School of Health and Rehabilitation Sciences and the National Institutes of Health through Grant Numbers UL1TR001857 and U24TR004111 (SV, NC, BM, JCS, MJB, YW). The Neptune data warehouse is supported by the National Center for Advancing Translational Sciences, The Center for Disease Control and National Institute for Occupational Safety and Health, NIH Common Fund and The Patient Centered Outcome Research Institute Clinical Research Network (UL1TR001857, RI-CRN-2020-006, OT2OD026554 and 5U24 OH009077 – to MJB, JCS, SV). The funders had no role in the design of the study, and collection, analysis, and interpretation of data and in preparation of the manuscript. The views presented in this report are not necessarily representative of the funder's views and belong solely to the authors.

**Statement on conflicts of interest**

The authors declare that they have no conflicts of interest.

**Summary table**

> - We developed a clinical research datamart, called ReDWINE (Rehabilitation Datamart With Informatics iNfrastructure for rEsearch), that transforms the rehabilitation-related electronic health record data collected from the a local health care system to the Observational Health Data Sciences and Informatics (OHDSI) Observational Medical Outcomes Partnership (OMOP) Common Data Model (CDM) to facilitate rehabilitation research.

- ReDWINE includes deployment of data visualization and data analytics tools to facilitate cohort definition and clinical data analysis.
- ReDWINE provide text analytical capabilities at scale by implementing a high-throughput natural language processing pipeline, the Open Health Natural Language Processing (OHNLP) toolkit.
- The data quality assurance analysis indicates the electronic health record data were successfully transferred to the OMOP CDM in ReDWINE.
- Two case studies were demonstrated the use of ReDWINE for rehabilitation research.

**Supplemental File**

**A. Details about the implementation for each schema in ReDWINE and the extract, transform, load (ETL) process.**

RAW Schema:

In the ReDWINE implementation, the RAW schema contains the following tables:

- Demographics: contains information such as patient identifier, birth date, death date, gender, race, and ethnicity.
- Diagnoses: contains information such as patient identifier, diagnosis type, diagnosis code (i.e., ICD code), diagnosis name, and diagnosis date;
- Encounters: contains patient medical encounter data, such as encounter identifier, encounter type, and encounter date;
- Encounter Notes: contains encounter notes scrubbed via NLMscrubber to best-effort de-identification (still containing some PHI) and related encounter metadata (patient identifier, encounter identifier, encounter date, etc.);
- Procedures: contains data related to medical procedures, such as procedure identifier, procedure type, procedure code;
- Procedure notes: contains procedure notes scrubbed via NLMscrubber to best-effort de-identification (still containing some PHI) and related procedure metadata (patient identifier, procedure identfier, procedure date).

We created these tables in the RAW schema and then loaded the EHR data we received from R3 into the tables using SQL programming language[1]. For structured EHR data, we used the original data without processing. For unstructured EHR data (encounter notes and procedure notes), an encounter or a procedure could have multiple entries of text data in the original format. Therefore, we ran a SQL preprocessing script to merge free text entries for each encounter and procedure. This turned out to be very helpful for the clinical NLP tool that we used for automatically extracting concepts from the notes and mapping them to standardized concept identifiers . We will describe this tool in detail in the next section.

---
[1] https://www.postgresql.org/docs/current/sql.html

CDM Schema:

The CDM schema contains data from several sources, including the EHR data obtained from the RAW Schema, OMOP's standardized vocabularies, and related metadata. This schema is generated using the CDM R package provided by the OHDSI community OHDSI/CommonDataModel[2]. The process, as shown in Figure A1, includes three steps: 1) generate the Data Definition Language (DDL) SQL programs using the tool; 2) make further changes to the DDL programs in an automated manner; and 3) run these programs inside ReDWINE.

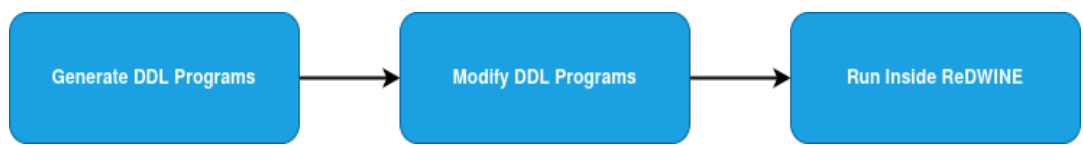

*Figure A1. The DDL Generation and Running Process*

This CDM R package allows for generating OMOP CDM DDLs of various versions. The one we used is CDM version 5.3. The reason is that all of the primary OHDSI tools support this CDM version. Depending on data, the DDL script might need further changes. In our case, we had to increase the maximum allowed length for text documents and change the integer types. The text document size change was needed since some of our documents were larger than the default maximum. As for integer type changes, we had to alter the DDL programs as the default integer type could not hold large numbers we used for representing person identifiers and note identifiers. We automated these DLL program transformations by defining all of these replacements in advance and incorporating them into an R program. Once the steps successfully finish, the DDL programs are run inside ReDWINE to create the cdm schema and all of the cdm schema tables as defined by the OMOP CDM.

---

[2] https://github.com/OHDSI/CommonDataModel

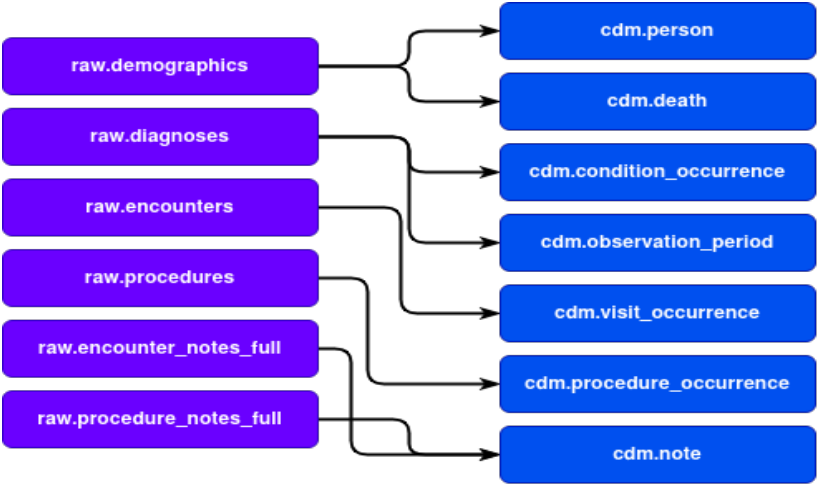

*Figure A2.Table Mapping from the raw Schema to the cdm Schema*

The ETL process is the most important process in the datamart and it should be performed with great care. Figure A3 depicts the piple for the ETL process. The CDM schema contains tables that are populated during the ETL process. During the ETL process, we ingest the data from the RAW schema, transform it, and load it into the CDM schema. For our pilot dataset, we leveraged a direct mapping from the RAW schema to the CDM schema, as shown in Figure A2. This mapping is not necessarily one-to-one mappings, meaning some tables (e.g., demographics) can be used to populate two different CDM tables. The mapping or data loading process virtually always involves some manual work in order to translate the data-specific concept language into the OMOP CDM compliant one. But once this work is done, the rest can be automated.

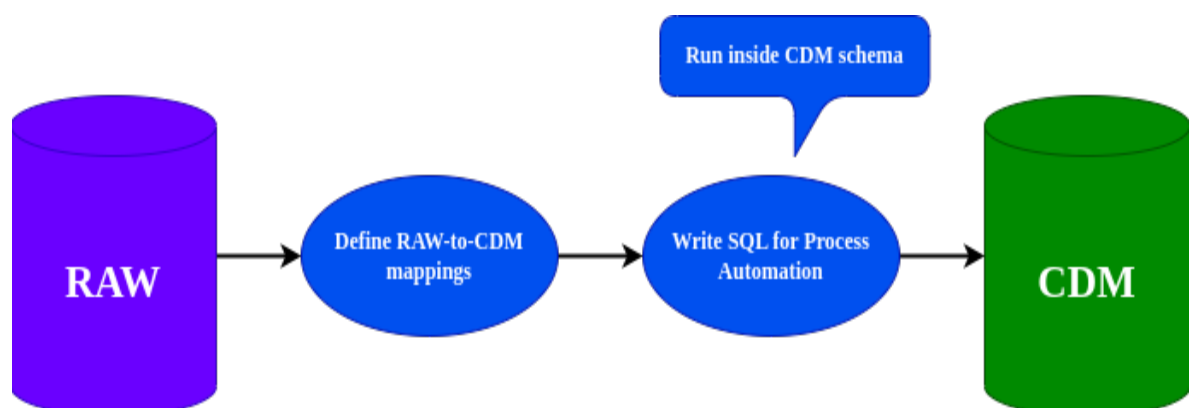

*Figure A3.The ETL Pipeline.*

The loading process first involves mapping the data in the RAW schema to the OMOP CDM supported standardized vocabulary. Automated mapping and manual mapping are two approaches that are usually adopted in this step. Automated mapping can be used for the code that follows widely used standards in healthcare domains, such as ICD-9/10, CPT-4, etc. In the automated mapping step, we leveraged the OMOP standardized vocabularies requested and obtained from OHDSI Athena[3] (note that the vocabularies need to be approved before downloading). We first loaded the OMOP vocabularies into the CDM schema and then applied the SQL function that we have developed to perform automated code-to-code mapping. The function utilizes the dictionaries that are loaded into the database and maps a code from the source data to the target CDM concept identifier. In our case, the source codes were ICD-9-CM, ICD-10, ICD-10-CM for the diagnosis, and CPT-4 for the procedure. An example of such a function call would look as follows: to_cdm(cpt_code, 'CPT4'), where the source code is from the CPT vocabulary and the target would be OMOP CDM concept ID. The SQL function we developed also allows for multiple-vocabulary lookups in which case it attempts to find if the source concept ID is present in one of the provided vocabularies. One example would be to_cdm(cpt_code, 'CPT4,ICD10'), when we are not sure whether the source code is in the CPT-4 vocabulary or ICD-

---

[3] https://athena.ohdsi.org/

10 vocabulary. In this case the function would first check if the code comes from the CPT-4 vocabulary and in case of a failed lookup, it would then move onto ICD-10 vocabulary and return the corresponding OMOP CDM concept identifier.

The manual mapping was performed with assistance from domain experts in rehabilitation when there is no mapping from the source concept to the OMOP CDM standardized vocabulary. This can be the case if the healthcare system has its own naming conventions. In practice, manual mapping in SQL is done using SQL's CASE statement[4] or similar to consider different cases and for every case, map it to the OMOP CDM standardized concept identifier. For example, in the ReDWINE implementation, we had to create mappings between the ethnicity concept (e.g., "HISPANIC OR LATINO" and "NOT HISPANIC OR LATINO") and the corresponding OMOP CDM concept identifiers (38003563 and 38003564, respectively). An example of the SQL code responsible for handling this mapping can be found in Table A1. A more complex case is mapping encounter visit types to the standardized vocabulary since all encounter visits are specific to the local health care system and clinical setting, sometimes with location names. Domain experts assisted in this process to look up the semantically closest concept in the OMOP CDM standardized vocabulary. If no similar concepts could be identified, we assigned a concept identifier of 0 per OHDSI's suggestion. Table A2 lists an example of SQL code for mapping enc_type column from RAW schema to OMOP CDM concept IDs in the CDM schema.

Table A1. An example of SQL code for handling ethnicity mapping.

```
1  CASE
2      WHEN ethnicity = 'HISPANIC OR LATINO'     THEN 38003563
3      WHEN ethnicity = 'NOT HISPANIC OR LATINO' THEN 38003564
4      ELSE 0
5  END AS ethnicity_concept_id
```

---

[4] https://www.postgresql.org/docs/current/functions-conditional.html

Once the standardized data were loaded, we performed a set of analyses using OHDSI's ACHILLES tool to ensure data quality. ACHILLES can be run on the CDM schema in order to generate descriptive statistics on the OMOP CDM database. These statistics can then be used by data analytics tools such as ATLAS[5]. The statistics themselves are loaded into the RESULTS schema.

Table A2. An example of SQL code for mapping enc_type column from RAW schema to OMOP CDM concept IDs in the CDM schema.

```
1   CASE
2       WHEN enc_type = 'ABSTRACT'              THEN 45877039
3       WHEN enc_type = 'ALLERGY MIXING'        THEN 44791052
4       WHEN enc_type = 'ALLERGY SHOT'          THEN 46272888
5       WHEN enc_type = 'AMBULATORY VISIT SUMMARY'  THEN 46237886
6       WHEN enc_type = 'ANTICOAGULATION'       THEN 35803400
7       WHEN enc_type = 'ANTICOAGULATION SCHEDULED' THEN 35803400
8       WHEN enc_type = 'APPOINTMENT'           THEN 4089197
9       --over 100 WHEN statements omitted
10      ELSE 0
11  END AS visit_concept_id
```

*Text Analytical Capabilities*

After performing ETL process followed by running the ACHILLES tool on the data, we used the clinical NLP tool to automatically extract medical concepts from encounter notes and procedure notes, and mapped them to the appropriate OMOP CDM standardized concept identifiers. We used the OHNLP Toolkit, which was developed by the OHNLP consortium, to automate concept extraction from the clinical notes. The OHNLP Toolkit uses is a ontology-derived dictionary-based approach and uses MedTagger [1] as the backbone for information extraction. Specifically, medical concepts were extracted from notes, normalized to Unified Medical Language System (UMLS) concept unique identifiers (CUIs), then mapped to associated OHDSI standardized Concept IDs, and finally stored in the Note_NLP table. An exact string matching algorithm was used to automate the mapping process and manual linking was created if an exact text match could not be found. Another benefit of using the OHNLP Toolkit is that one can define customized rules for extracting specific concepts and phenotypes from notes. The process will populate the Note_NLP table in the CDM schema.

---

[5] https://github.com/OHDSI/Atlas

The OHNLP Toolkit uses Apache Flink, an open-source, unified stream-processing and batch-processing framework, to observe the batch text data processing process and corresponding statistics, as shown in Figure A4. We found this feature to be extremely helpful for debugging . Specifically, for ReDWINE, the initial "Task Heap" memory was not large enough, which would result in unsuccessful exits. The tool helped us debug the issue and we then increased the memory to 52.0 GB. The OHNLP Toolkit took around 8 hours to complete the text analysis for over 2 million notes and concept extraction and populated these concepts in the Note_NLP table of the CDM schema. More details about the OHNLP Toolkit can be found in the previous study [2].

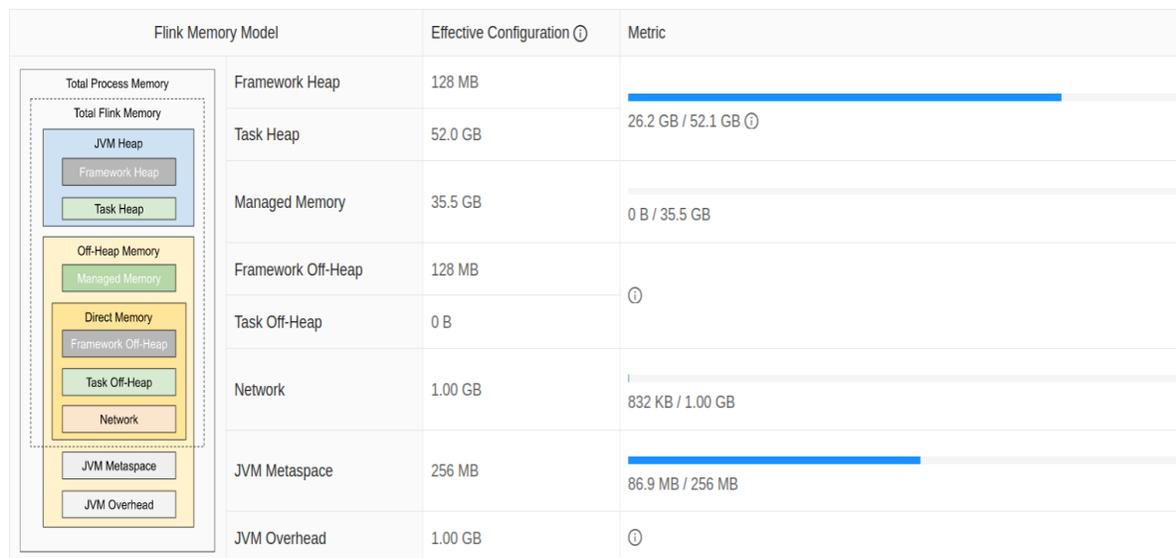

*Figure A4.Apache Flink Running on the Server via OHNLP Toolkit.*

*Data Analytic Tools*

Broadsea is a tool that allows for running the core OHDSI data analytics stack inside a Docker container[6]. This includes the ATLAS web application and the HADES R packages within a RStudio Server. The containerization of the core stack allows for the cross-platform implementation. This means that ReDWINE can be deployed on most major platforms including Linux, macOS, and Windows. In our case, we used Linux for both hosting our database as well as deploying Docker containers. It should be noted that we migrated to Broadsea 2.0 a few weeks after the release since the broad support from the OHDSI community. For the most part, it was a painless process, but it does involve configuration of the database connections for Broadsea 2.0 to access the CDM data as well as the ACHILLES-populated results table. After deploying the

---

[6] https://www.docker.com/resources/what-container/

Docker containers, we have both ATLAS as well as HADES running inside our containers and researchers can access both through the Broadsea 2.0 portal running on ReDWINE.